\documentclass[runningheads]{llncs}
\usepackage[utf8]{inputenc}
\usepackage[T1]{fontenc}
\usepackage{graphicx} 
\usepackage{booktabs}
\usepackage{amsfonts}
\usepackage{multirow}
\usepackage{amsmath}
\usepackage{hyperref}
\usepackage[sort]{cite}
\usepackage{authblk} 
\usepackage{orcidlink} 

\author{Cedric Möller\inst{1}\orcidID{0000-0001-6700-3482} \and
Ricardo Usbeck\inst{2}\orcidID{0000-0002-0191-7211}}
\authorrunning{C. Möller and R. Usbeck}
\institute{Universität Hamburg, Department of Informatics, Semantic Systems, Germany \email{cedric.moeller@uni-hamburg.de} \and
Leuphana Universität Lüneburg, Institute for Information Systems, Artificial Intelligence and Explainability, Germany \email{ricardo.usbeck@leuphana.de}\\
}
\title{Analyzing the Influence of Knowledge Graph Information on Relation Extraction}

\begin{document}

\maketitle

\begin{center}
  \fbox{
    \parbox{0.9\linewidth}{
      \textbf{Preprint Notice:} This is a preprint of the following paper:  
      Cedric Möller and Ricardo Usbeck, "Analyzing the Influence of Knowledge Graph Information on Relation Extraction"  
      in "Lecture Notes in Computer Science", vol. 15718 (The Semantic Web – ESWC 2025), pp. 460–480.  
      First online published on June 1, 2025.  

      The final authenticated version is available at:  
      
      \url{https://link.springer.com/chapter/10.1007/978-3-031-94575-5_25}
    }
  }
\end{center}

\begin{abstract}
    We examine the impact of incorporating knowledge graph information on the performance of relation extraction models across a range of datasets. Our hypothesis is that the positions of entities within a knowledge graph provide important insights for relation extraction tasks. We conduct experiments on multiple datasets, each varying in the number of relations, training examples, and underlying knowledge graphs. Our results demonstrate that integrating knowledge graph information significantly enhances performance, especially when dealing with an imbalance in the number of training examples for each relation.
    We evaluate the contribution of knowledge graph-based features by combining established relation extraction methods with graph-aware Neural Bellman-Ford networks. These features are tested in both supervised and zero-shot settings, demonstrating consistent performance improvements across various datasets.
\end{abstract}

\section{Introduction}
Populating an existing knowledge graph (KG) with new information is an essential challenge. One subtask integral for this is relation extraction.
Relation extraction is the task of identifying the expressed relation between two entities. The focus of this usually resides on a relation expressed in a sentence, document or between multiple documents. 
In contrast to that, link prediction~\cite{DBLP:journals/symmetry/WangQW21} infers potential relations based on the structure of a knowledge graph~\cite{DBLP:journals/tnn/JiPCMY22}.

In this paper, our goal is to combine both ways of tackling the task, based on the text and based on the graph, in a single framework. For that, we incorporate a Neural Bellman Ford (NBF) network~\cite{DBLP:conf/nips/ZhuZXT21} into the relation extraction process, allowing us to include graph-based information while being generalizable to new entities in the knowledge graph. This leads to a model that considers information in the KG and in the document jointly. 

We study the impact of the KG data on several established datasets, with a focus on both supervised and zero-shot scenarios.  To tackle the supervised and zero-shot scenarios, we use two different versions of the NBF network, one which assumes that all encountered relations in the graph are known before and another that creates the relation representations on the fly. 
As we focus solely on relation extraction, we assume that the actual entity mentions in the text are pre-annotated. Therefore, the task is to determine whether a relationship exists between a pair of entities and, if so, identify which specific relationship holds.

Unlike existing methods~\cite{Jain2024,Bastos2021}, our approach does not rely on learned entity embeddings, allowing it to generalize to new entities. Additionally, we introduce a method that is also suitable for zero-shot settings.

Our contributions are an analysis of the impact of knowledge graph information in the following relation extraction settings:

1. \textbf{Supervised Setting:} We investigate how incorporating knowledge graph information enhances model performance. This includes improvements in accuracy by providing richer contextual information and better resolving ambiguities.

2. \textbf{Zero-shot Setting:} We explore the role of knowledge graphs in zero-shot relation extraction, demonstrating their potential to enable models to generalize to unseen categories by providing semantic background and auxiliary data for informed predictions.

\paragraph{Code:} Our code is available at  \\\url{https://github.com/semantic-systems/kg-based-re}.

\section{Method}

\subsection{Problem Definition}
We target the relation extraction problem. Given a text and a pair of annotated entities within the text, the goal is to identify the relation expressed between the two entities out of the set of all relations $R$. Depending on the input text, multiple entities might be mentioned in the text. For each pair of entities multiple relations are potentially expressed. We refer to this as document-level relation extraction. If there are only two entities between which the relation needs to be extracted, we refer to it to sentence-level relation extraction.

In a \textbf{supervised setting}, the set of potential relations is known during training. It is possible that no relation is expressed between two entities, which can also be interpreted as an additional \texttt{no\_relation} relation. 
When one considers the \textbf{zero-shot setting}, the assumption is that the set of relations seen during training differs from the set of relations encountered during evaluation. To solve this task, the method has to generalize to unseen relations.

In addition to textual information, we assume the availability of a knowledge graph . A knowledge graph is a graph consisting of nodes and edges. Each node corresponds to a specific entity, either depicting some individual or concept. The edges between nodes are equipped with a relation that denotes some relationship between two nodes. Two nodes linked by a relation are commonly also denoted as a triple. 
Formally, we define the graph as \( G = (V, E) \), where \( V \) is the set of nodes and \( E \subseteq V \times V \times R_G \) is the set of edges with $R_G$ being the edges existing in the graph. $R_G$ does not need to overlap with $R$. 
For example, an edge equipped with the \texttt{married\_to} relation between the subject \texttt{Barack Obama} and the object \texttt{Michele Obama} expresses that both those persons are married to each other. Multiple edges may hold between two nodes. In this work, we assume that the corresponding nodes of the entities marked in the text are known and can be used to incorporate KG-internal information.

\begin{figure}
    \centering
    \includegraphics[width=0.9\linewidth]{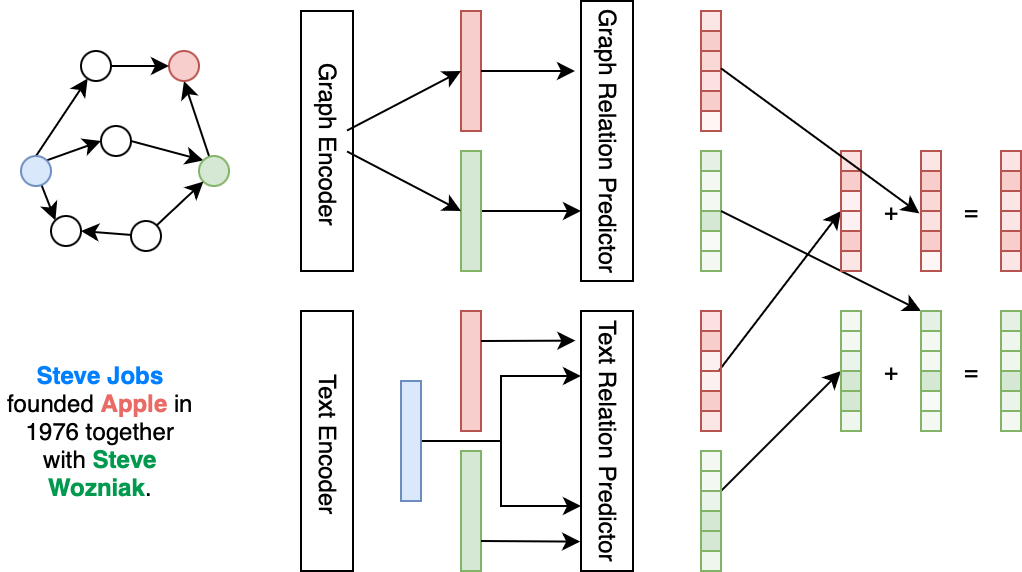}
    \caption{Model architecture: The figure illustrates relation prediction between a subject (blue) and two objects (red and green). Text and graph are encoded to predict relations involving Steve Jobs. The graph predictor identifies likely graph-based relations, while the text predictor identifies text-expressed relations. Both scores are combined. Identically colored mentions and nodes represent the same entity, and the predictors output a relation distribution, operating in either supervised or zero-shot modes. }
    \label{fig:model}
\end{figure}

Our method consists of two main components (see Figure~\ref{fig:model}), the textual module and the graph module. 
The textual module purely works on the textual input, while the graph module considers the potential relation between two mentioned entities as expressed in the graph. 
\subsection{Textual Module}
\paragraph{Supervised.}
The input text consists of the marked entities where each entity mention is surrounded by \$-signs.
The textual encoder follows a well-established architecture as proposed by Zhou et al.~\cite{Zhou2021}. Initially, the textual input is encoded using an encoder-only model 
$$H = \text{Encoder}(X), \quad H = [h_1, h_2, \dots, h_N], \quad h_i \in \mathbb{R}^d .$$
Then, for each of the marked entity mentions $m_j$, the encoded token of its left-side \$-sign is taken, denoted as $h_{m_j} = h_{p_j}$ where $p_j$ is the position of the sign. Additionally, the attention scores for the \$ to all other tokens throughout all layers are averaged over all layers and normalized 
$$\bar{a}_{p_j} = \frac{1}{L} \sum_{\ell=1}^L a^{(\ell)}_{p_j}$$ with $a^{(\ell)}_{p_j} = A^{(\ell)}[p_j, :]$.
As multiple entity mentions $M_\text{entity}$ might exist in a document for a single $entity$, the attention scores are averaged over all entity-specific mentions 
$$\tilde{a}_{\text{entity}} = \frac{1}{|M_\text{entity}|} \sum_{j=1}^{|M|} \bar{a}_{p_j}$$ and the \$-sign encodings are pooled by applying the \texttt{logsumexp} operation  to get the entity-encoding
$$\tilde{h}_{\text{entity}} = \texttt{logsumexp}(\{h_{m_1}, h_{m_2}, \dots, h_{m_{|M_\text{entity}|}}\}).$$

This gives us two key representations per entity: the attention $\tilde{a}_{\text{entity}}$ to all other tokens and the entity's \$-encoding $\tilde{h}_{\text{entity}}$, which are used in the next steps. For each pair of entities, we compute a combined representation by first point-wise multiplying the attention scores of both entities $\tilde{a}_k \odot \tilde{a}_l$ and normalizing them giving $\tilde{a}_{k,l}$. Then, we compute an attention-based pair representation by performing a weighted sum over all encoded tokens, resulting in  
$$c = \sum_{i=1}^N \tilde{a}_{k,l}[i] \cdot h_i.$$

Finally, the subject encoding is calculated by 
$$s_k = W_s \text{concat}\left(\tilde{h}_{_k}, c\right) + b_s$$ where $W_s$ is a weight matrix and $b_s$ a bias.
The same is done to get an object encoding $o_l$.
Both are used to predict the scores for all relations by applying a bilinear mapping $$p_t = s_k^{\top} W o_l \in \mathbb{R}^R$$ where $W\in \mathbb{R}^{d \times R \times d}$ and $R$ is the number of relations.

\paragraph{Zero-shot.}
For zero-shot learning, we employ a relation extraction approach via multiple-choice classification~\cite{Lan2023}. For each document \( s \), we concatenate it with the label \( l_r \) and description \( d_r \) of a relation \( r \)
\[
x_r = \text{concat}(s, l_r, d_r).
\]

The concatenated input \( x_r \) is then fed into an encoder-only model to obtain the token encoding
\[
H_r = \text{Encoder}(x_r), \quad H_r = [h^{(r)}_1, h^{(r)}_2, \dots, h^{(r)}_N].
\]

We extract the encoding of the first token \( h^{(r)}_1 \) and project it using a two-layer mapping to a score
\[
\text{score}(r) = W_2 \, \sigma(W_1 h^{(r)}_1 + b_1) + b_2
\]
where \( W_1 \) and \( W_2 \) are weight matrices, \( b_1 \) and \( b_2 \) are biases, and \( \sigma \) is the tanh function.

This process is repeated for all relations \( r \in R \) to compute their respective predictions, giving $p_t \in \mathbb{R}^R$.

\subsection{Graph Module}
\paragraph{Supervised.}
For the graph-encoder, we follow a modified version of the Neural Bellmann-Ford (NBF) graph neural network~\cite{DBLP:conf/nips/ZhuZXT21}. Originally designed for link prediction, this method takes a subject entity and a relation as input. Given the large number of relations in our dataset, running this model for each relation introduces significant computational overhead, which was not feasible with our available resources.  

To address this, we eliminate the need for specifying an input relation. Instead, we predict all possible output relations based on the final representation of each node.  

Initially, each node in the graph is initialized with a zero vector, except for a designated start node, which is assigned a specific vector \( g_{\text{start}} \):  
\[
g_v^{(0)} = \begin{cases}  
g_{\text{start}}, & \text{if } v = v_{\text{start}}, \\  
\mathbf{0}, & \text{otherwise}.  
\end{cases}
\]  

The graph undergoes \( T \) iterations of message passing, where each edge \( (u, v) \) is represented by its direction and associated relation \( r \). The message passing consists of three main steps for each node \( v \):  

1. Propagation: Information is propagated along the edges using the DistMult operation~\cite{DBLP:journals/corr/YangYHGD14a} 
\[
m_{u \to v}^{(t)} = g_u^{(t-1)} \odot r_{(u,v)}
\]  
where \( m_{u \to v}^{(t)} \) is the message from node \( u \) to node \( v \), $r_{(u,v)}$ is a relation-specific representation and \( \odot \) denotes element-wise multiplication.  

2. Aggregation: Incoming messages are aggregated for each node \( v \) using Principal Neighbourhood Aggregation (PNA)~\cite{DBLP:conf/nips/CorsoCBLV20} 
\[
m_v^{(t)} = \text{Aggregate}\left(\{m_{u \to v}^{(t)} \mid u \in \mathcal{N}(v)\}\right)
\]  
where \( \mathcal{N}(v) \) denotes the neighbors of \( v \).  

3. Update: The node's representation is updated via a linear projection and a non-linear activation with  
\[
g_v^{(t)} =  \mu(W^{(t)} m_v^{(t)} + b^{(t)})
\]  
where \( W^{(t)} \) and \( b^{(t)} \) are the weights and biases of the linear transformation at iteration \( t \) and \( \mu \) is the ReLU activation function.  

After \( T \) iterations, the message passing is stopped, and the representation of each node is retrieved. This representation captures the relation of any node with respect to the start node.  

In our case, we initialize the start node as the node corresponding to the subject entity and retrieve the representations of each object entity node $g_l$. This representation \( g_{l} \) is then passed through a two layer network to predict the scores for all relations
\[
p_g = W_4 \, \mu(W_3 g_{l} + b_3) + b_4 \in \mathbb{R}^R
\]  
where \( W_3 \), \( W_4 \), \( b_3 \), and \( b_4 \) are weights and biases.

\paragraph{Zero-shot.} For zero-shot learning, we rely on the zero-shot variant of the NBF network, denoted as ULTRA~\cite{DBLP:conf/iclr/0001YM0Z24}. ULTRA consists of two components: a relation graph encoder and an entity graph encoder, both implemented as NBF networks.  

The relation graph is constructed with nodes representing relations and edges connecting them if the subject/object of one relation is the subject/object of another.  
A designated relation is set as the start, and the relation graph encoder produces a representation for each relation node \( h_r \)  
\[
h_{r} = \text{RelationGraphEncoder}(r_{\text{start}})
\] where $r \in R_G$.

These relation representations \( h_r \) are used in the entity graph encoder, which outputs entity representations \( h_{l} \) conditioned on a start node and the relation. These are then fed into a multi-layer projection to compute a single score
\[
\text{score}(r) = W_6 \, \mu(W_5 h_{\text{entity}} + b_5) + b_6
\]  
where \( W_5 \), \( W_6 \), \( b_5 \), and \( b_6 \) are weights and biases, and \( \mu \) is the ReLU activation function.  

To classify all relations, ULTRA is run for each relation in the set $R$, and the predictions are concatenated:  
\[
p_g = [\text{score}(r_1), \text{score}(r_2), \dots, \text{score}(r_R)] \in \mathbb{R}^R
\]

\subsection{Final Prediction}
To compute the final prediction, we integrate the logits from both sources by accumulating them. Let \( \alpha \) and \( \beta \) represent the respective weights for each source. The final logits are computed as
\[
p = p_t + p_g \in \mathbb{R}^R
\]
where \( p_t \in \mathbb{R}^R \) and \( p_g \in \mathbb{R}^R \) are the logits from the two predictive models (either for the supervised or zero-shot setting), and \( R \) represents the number of relations.  

\subsection{Post-Prediction}
In document-level relation extraction, rule-learning techniques are frequently employed to enhance the predictive capability of models~\cite{Ru2021,Fan2022,DBLP:conf/acl/QiDW24}. These approaches leverage an initial set of relations generated by a text-only model and perform reasoning using learned rules to infer additional relations.  

We adopt a similar methodology for document-level relation extraction tasks. Specifically, starting with the initial predictions produced by the textual component, we enrich the underlying graph. This enrichment involves adding new edges to the graph between each pair of nodes \( (v_i, v_j) \) whenever a relation is identified between the corresponding entities in the input text.  

 \( R_G \) denote the set of predefined relation types in the initial graph. To distinguish between a priori known relations and those inferred through reasoning, we define an additional set of relation types \( R' \) corresponding exclusively to the predicted relations. The graph \( G = (V, E) \) is extended by introducing new edges:  
\[
E' = \{(v_i, v_j, r') \mid r' \in R', v_i, v_j \in V, \text{ and } r' \text{ is predicted for } (v_i, v_j)\}.
\]  
The extended graph is then represented as \( G' = (V, E \cup E') \).  

\begin{figure}
    \centering
    \includegraphics[width=0.9\linewidth]{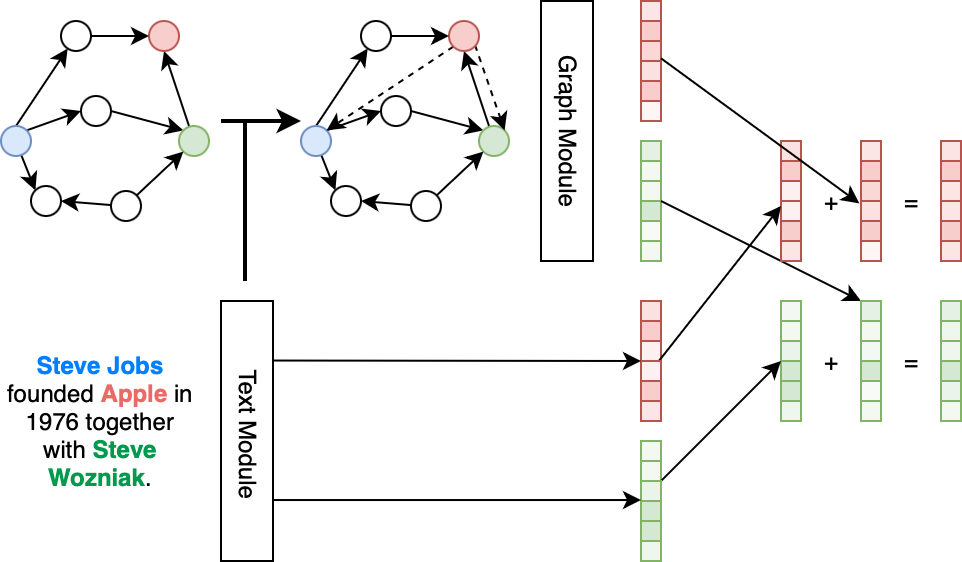}
    \caption{The figure illustrates the post-prediction mechanism. The input text is initially processed through the textual module to identify relations. These identified relations are then incorporated into the input graph for the graph module. Using the updated graph and the initial textual predictions, the final predictions are generated. }
    \label{fig:pp}
\end{figure}

By explicitly encoding these predicted relations as a separate set \( R' \), the model can effectively distinguish between the original relations \( R \) and the newly inferred ones, thus enabling more nuanced reasoning and relation classification.  Post-Prediction is only used for document-level relation extraction. See Figure~\ref{fig:pp} for an overview.

\subsection{Losses}
For datasets with more than two entities per example, we use the HingeABL loss~\cite{DBLP:conf/emnlp/WangLPC23}, which is designed to more-gracefully handle the problem of imbalance in multi-class classification optimization in contrast to a regular binary cross-entropy loss. For sentence-level relation extraction datasets, we use the cross-entropy loss to optimize the models.

\section{Evaluation}
\subsection{Setup}
We evaluate the influence of the graph information on several datasets. As the encoder models we used RoBERTa-large for Re-DocRED, BERT-base for DWIE and BioBERT for BioRel. For Wiki-ZSL and FewRel, we relied on BERT-base. We chose those to be comparable to other existing methods.

\paragraph{Supervised datasets.} As supervised datasets, we use the Re-DocRED~\cite{Tan2022a}, DWIE \cite{Zaporojets2021} and BioRel~\cite{DBLP:journals/bmcbi/XingLS20} dataset. We chose those datasets as they are available with annotated entity mentions and were evaluated on by past methods that utilised KG information. An overview of them can be found in Table~\ref{tab:dataset_statistics_sup}.

For Re-DocRED and DWIE, we rely on Wikidata as the corresponding knowledge graph. For BioRel, we rely on two available ontologies: Medline~\cite{DBLP:conf/prima/Yang03} and NCIt~\cite{DBLP:conf/aime/KumarS05}. 

We gathered each ontology and linked the entity mentions to the corresponding nodes. On Re-DocRED, 83\%  of all mentions were linked to a Wikidata node. On DWIE, this only applies to 53\% of all mentions. The rest were either literals or were not linked during dataset creation. On BioRel, 82\% of all entity mentions were covered. The others were not identifiable in the two available ontologies.
DWIE and Re-DocRED are both document-level relation extraction datasets with DWIE containing considerably longer documents than Re-DocRED. BioREL is a sentence-level relation extraction dataset. 

As the main metric, we use F1-measure. For Re-DocRED and DWIE, we use the micro F1-measure, where true positives correspond to correctly classified relations, false negatives to relations which are classified between entities where no relation is expressed, and false positives correspond to incorrectly predicted relations (even if a relation is expressed between the two entities). Furthermore, for Re-DocRED, we report Ign-F1 which is the F1 while filtering out all triples that already occur during training.

For BioRel, we measure the Macro-F1 which is the averaged F1 over all relation classes.

For all datasets, we train each method three times and report the averaged metrics.

\begin{table}[h]
    \centering
    \begin{tabular}{lcccc}
        \toprule
        \small \textbf{Dataset Name} & \small \textbf{\# Documents} & \small \textbf{\# Relations} & \small \textbf{\# Mentions} & \small \textbf{\# Triples} \\
        \midrule
        DWIE & 802 & 65 & 43,373 & 317,204 \\
        Re-DocRED & 4053 & 97 & 132,375 & 120,539 \\
        BioRel & 533,560 & 125 & 1,067,120 & 533,560 \\
        \bottomrule
    \end{tabular}
    \caption{Supervised relation extraction datasets}
    \label{tab:dataset_statistics_sup}
\end{table}

\paragraph{Zero-shot datasets.} For the zero-shot evaluations we rely on the popular FewRel \cite{Han2018,Chia2022} and Wiki-ZSL~\cite{Chen2021} datasets (see Table~\ref{tab:dataset_statistics_zero} for an overview). These are sentence-level relation extraction datasets. Both contain annotated entity mentions linked to Wikidata~\cite{DBLP:journals/cacm/VrandecicK14}. The datasets are prepared in three versions, with 5, 10 or 15 test relations. Furthermore, for each of the versions, the full dataset is resampled five times. That means, the model is evaluated on five different runs for each version to compensate for the high variance due to the small number of test relations. 
All mentioned entities are linked to Wikidata.
Macro F1-measure is calculated to evaluate the performance of methods.

\begin{table}[h]
    \centering
    \begin{tabular}{lcccc}
        \toprule
        \small \textbf{Dataset Name} & \small \textbf{\# Documents} & \small \textbf{\# Relations} & \small \textbf{\# Mentions} & \small \textbf{\# Triples} \\
        \midrule
        FewRel & 70,000 & 100 & 140,000 & 70,000 \\
        Wiki-ZSL & 94,383 & 113 & 132,375 & 188,766 \\
        \bottomrule
    \end{tabular}
    \caption{Zero-shot relation extraction datasets}
    \label{tab:dataset_statistics_zero}
\end{table}

\paragraph{Graph.} For each entity in a document, we gather its two-hop neighborhood in the knowledge graph, limiting outgoing and incoming edges to 100 per hop. This results in a maximum of 10,000 entities per entity's neighborhood. By doing this for each entity, the model can explore up to four-hop connections between entities. Additionally, we remove nodes that appear in only a single triple unless they are part of the original set of entities, which helps manage large datasets with up to 200,000 nodes per subgraph.

For sentence-level relation extraction, we eliminate direct triples between subjects and objects to prevent trivial relations, especially in distantly-supervised datasets.~\footnote{Distant-supervision means that the documents were annotated using the information in the underlying KG. By not removing them, the model might simply check whether the relation already holds in the graph, making the task trivial.} In the zero-shot setting, we sample 1000 triples per relation, extract the two-hop neighborhood for each subject and object, and construct the relation graph. We filter noisy connections by retaining only relation-relation edges present in at least 10\% of the sampled neighborhoods.

The graph neural networks are run with four layers of message passing to fully utilise the maximum-hop-distance between two individual entities. We experimented with more hops but did not achieve any improvements in performance.

\paragraph{Methods.}
In our analysis on DWIE, we evaluate a range of document-level relation extraction methods, referencing works such as \cite{Xu2021,Vashishth2018,Verlinden2021}. We specifically focus on comparing the performance of the top models utilizing rule-learning techniques, as highlighted in \cite{Ru2021,Fan2022,DBLP:conf/acl/QiDW24}. Additionally, we incorporate two methods that integrate knowledge graph information, as explored by \cite{Bastos2021,Jain2024}.

On ReDocRED, we compare against several BERT-based and RoBERTa-based methods~\cite{Zhou2021,Xu2021,Tan2022,Zhang2021,DBLP:conf/eacl/MaWO23} again including two that incorporate knowledge graph information~\cite{Jain2024,Bastos2021} as well.

On BioREL we compare against the best-performing non-biomedical \cite{Bastos2021,Sorokin2017,DBLP:conf/acl/ZhuLLFCS19,DBLP:journals/jmlr/RaffelSRLNMZLL20} and biomedical relation extraction methods~\cite{Jain2023} as reported by Jain et al.~\cite{Jain2023}. 

On Wiki-ZSL and Fewrel we compare against several zero-shot methods, ranging from entailment-based methods~\cite{Rocktaeschel2016}, encoding-based methods \cite{Chen2021a,Tran2022,tran2023enhancing,Zhao2023}, generative methods~\cite{Chia2022} and discriminative prompting methods~\cite{Lv2023,Lan2023}

In our supervised experiments, we refer to our method as ATLOP-KG when incorporating the knowledge graph  component into the ATLOP architecture, and as ATLOP-KG-PP when including both the KG component and the post-prediction module. Additionally, we present our baseline results using the ATLOP architecture with the HingeABL loss, labeled as ATLOP-Hinge. Furthermore, we denote a method solely using the graph component as NBF.

For the zero-shot experiments, we designate our approach as MC-BERT-KG, and we also present results for MC-BERT with added descriptions, referred to as MC-BERT w/ descriptions.

\paragraph{Hyperparameters.}
All training and inference runs were performed on a single NVIDIA A6000 GPU. 
We used a batch size of $8$ for each supervised training run and a batch size of $16$ for the unsupervised runs. We use learning rates of $3e-5$ for the text encoders and of $1e-4$ for the graph encoders. We trained each method by relying on early stopping based on the validation F1-measure. 

\subsection{Results}

\paragraph{Supervised.}

 \begin{table}[htb!]
    \centering
    \begin{minipage}{.45\linewidth}
        \centering
        \begin{tabular}{lll}
        \toprule
        \textbf{Model} &\textbf{F1} \\ \midrule
        DRN*GloVe~\cite{Xu2021}  & 56.04 \\ 
        RESIDE~\cite{Vashishth2018} &66.78 \\ 
        RECON~\cite{Bastos2021}  &66.94 \\ 
        KB-Graph~\cite{Verlinden2021}  &66.89 \\ 
        ATLOP~\cite{Zhou2021} & 75.13 \\
        DocRE-CLiP~\cite{Jain2024} &67.10 \\ 
        LogicRE-ATLOP~\cite{Ru2021}  &75.67 \\ 
        MILR-ATLOP~\cite{Fan2022}  &76.51 \\ 
        JMLR-ATLOP~\cite{DBLP:conf/acl/QiDW24}  &77.85 \\ 
        ATLOP-Hinge (Ours) &77.35\\
        NBF (Ours) &44.27\\ 
        ATLOP-KG (Ours)&\underline{78.50}\\
        ATLOP-KG-PP (Ours)&\textbf{79.46}\\
        \bottomrule
        \end{tabular}
        \caption{F1 Scores on DWIE}
        \label{tab:dwie}
    \end{minipage}
    \hspace{0.5cm}  
    \begin{minipage}{.45\linewidth}
        \centering
        \begin{tabular}{lll}
\toprule
\textbf{Model} & \textbf{Ign-F1} &\textbf{F1} \\ \midrule
ATLOP~\cite{Zhou2021} & 76.82 & 77.56 \\ 
DRN*~\cite{Xu2021} & 74.3& 75.6 \\ 
KG-DocRE~\cite{Tan2022} & 80.32 & 81.04 \\ 
DocuNet~\cite{Zhang2021} & 78.52& 79.64 \\ 
DREEAM~\cite{DBLP:conf/eacl/MaWO23} & \underline{80.39}& \underline{81.44} \\ 
DocRE-CLiP~\cite{Jain2024} & \textbf{80.57}& \textbf{81.55} \\
ATLOP-Hinge (Ours) & 76.97 & 78.07\\
NBF (Ours) &46.94 & 49.58 \\
ATLOP-KG (Ours) & 77.70 & 78.70\\
ATLOP-KG-PP (Ours) & 77.80 & 78.83\\
\bottomrule
\end{tabular}
\caption{F1 Scores on ReDocRED}
\label{tab:rdocred}
    \end{minipage}
\end{table}

Our method outperforms recent state-of-the-art methods on the DWIE dataset by 1.5 F1-measure points, as illustrated in Table~\ref{tab:dwie}. Notably, even without graph information, our text-based relation extraction architecture performs comparably to existing state-of-the-art models. The inclusion of graph information further enhances performance, which is remarkable given our reliance on the same architectural framework as the state-of-the-art. While the inclusion of KG information already leads to a larger improvement, using the post-prediction step as well improves it even more. We hypothesize that our adaptations for handling long documents, typical in DWIE, contribute significantly to this improvement. This adaptation involves splitting input documents based on the encoder's maximum token length while introducing a stride to maintain contextual coherence. This approach seems to compensate for adjustments potentially overlooked in previous state-of-the-art methods. Without that, our performance diminishes to the one as reported in the paper by Qi et al.\cite{DBLP:conf/acl/QiDW24}.

\begin{table}[htb!]
\centering
\begin{tabular}{lll}
\toprule
\textbf{Model} & \textbf{Accuracy} & \textbf{Macro F1 Score} \\ \midrule
GPGNN~\cite{DBLP:conf/nips/ZhuZXT21} & 85&84.00 \\ 
ContextAware~\cite{Sorokin2017} &89 &87.00 \\ 
T5~\cite{DBLP:journals/jmlr/RaffelSRLNMZLL20} & 88&86.00 \\ 
RECON~\cite{Bastos2021} & 89.6 &86.00 \\ 
DocRE-CLiP~\cite{Jain2023}& 92 &90.00\\ 
NBF & 75.35 & 74.74\\ 
ATLOP-Hinge (Ours) & \underline{96.69} & \underline{95.71} \\
ATLOP-KG (Ours) & \textbf{97.93} &\textbf{96.90}\\\bottomrule
\end{tabular}
\caption{Accuracy and F1 Scores on BioREL}
\label{tab:biorel}
\end{table}

Conversely, our approach does not achieve state-of-the-art performance on the Re-DocRED dataset (Table~\ref{tab:rdocred}). Although incorporating knowledge graph (KG) information enhances performance over the text-only model, the post-prediction step contributes minimally. This outcome may stem from the typically smaller document sizes in Re-DocRED, reducing the advantage offered by the KG enhancements. Additionally, our lack of pre-training and evidence fusion, which distinguish the leading approaches~\cite{DBLP:conf/eacl/MaWO23}, might explain this discrepancy. Nevertheless, we chose to forego these complex stages to minimize computational overhead, focusing instead on exploring the fundamental effects of graph integration.
The unexpectedly high performance of DocRE-CLiP~\cite{Jain2024} is notable, particularly given their claims regarding the significant impact of integrating graph information. However, upon reviewing their paper and accompanying code, we were unable to ascertain the specific methods they employed to incorporate this information. Additionally, there is an indication that their training process encompasses entities from not only the training set but also the test and validation sets.

\begin{table}[!htb]
    \centering
    \begin{tabular}{cccccccc}
    \toprule
        & & \multicolumn{3}{c}{Wiki-ZSL} & \multicolumn{3}{c}{FewRel} \\
         $m$ & Model & P & R & F1 & P & R & F1  \\
         \midrule \multirow{8}{*}{5}& CIM~\cite{Rocktaeschel2016} & 49.63 & 48.81 & 49.22 & 58.05 & 61.92 & 59.92 \\
         &  ZS-BERT~\cite{Chen2021a} &71.54 &72.39 &71.96 &76.96 &78.86& 77.90 \\
         &  Tran et al. (2022)~\cite{Tran2022} &87.48 &77.50& 82.19& 87.11 &86.29 &86.69 \\
         &  RelationPrompt NG~\cite{Chia2022} & 51.78 &46.76& 48.93 &72.36& 58.61 &64.57 \\
         &  RelationPrompt~\cite{Chia2022} & 70.66& 83.75& 76.63 &90.15 &88.50& 89.30 \\
         & RE-Matching~\cite{Zhao2023} & 78.19 & 78.41  & 78.30 & 92.82 & 92.34 & 92.58 \\
         & DSP-ZRSC~\cite{Lv2023} & \underline{94.1} & 77.1 & 84.8 & 93.4 & 92.5 & 92.9 \\
         &  Tran et al. (2023)~\cite{tran2023enhancing} &\textbf{94.50} &\textbf{96.48}& \textbf{95.46}& \textbf{96.36} &\textbf{96.68} &\textbf{96.51} \\
         &  MC-BERT~\cite{Lan2023} & 80.28 &84.03& 82.11& 90.82 &90.13& 90.47 \\
         & MC-BERT w/ descriptions (Ours) & 85.00 & 84.41 & 84.68 & 93.33 & 92.50 & 92.91  \\
         &  MC-BERT-KG (Ours) & 88.89 & 89.46 & 88.92 & 88.39& 91.37  & 92.02 \\
         \midrule \multirow{8}{*}{10}&   CIM & 46.54 &47.90 &45.57 &47.39 &49.11& 48.23 \\
         &  ZS-BERT & 60.51& 60.98& 60.74 &56.92 &57.59& 57.25 \\
         &  Tran et al. (2022) & 71.59 &64.69& 67.94 &64.41& 62.61& 63.50 \\
         &  RelationPrompt NG & 54.87 &36.52 &43.80& 66.47& 48.28& 55.61 \\
         &  RelationPrompt & 68.51& 74.76& 71.50& 80.33 &79.62& 79.96 \\
         & RE-Matching & 74.39 & 73.54 & 73.96 & 83.21 & 82.64 & 82.93 \\
         & DSP-ZRSC & 80.0 & 74.0 & 76.9 & 80.7 & \textbf{88.0} & 84.2 \\
         &  Tran et al. (2023) &\textbf{85.43} &\textbf{88.14}& \textbf{86.74}& 81.13 &82.24 &81.68\\
         &  MC-BERT & 72.81 &73.96& 73.38& 86.57 &\underline{85.27} &85.92 \\
         & MC-BERT w/ descriptions (Ours) & 74.89 & 76.05 & 75.46 & 85.16 & 83.36 &  84.24\\
         &  MC-BERT-KG (Ours) & \underline{81.72} & \underline{80.52} & \underline{81.10} & \textbf{88.63}  & 84.80 & \textbf{86.63} \\
         \midrule \multirow{8}{*}{15}& CIM & 29.17& 30.58 &29.86& 31.83& 33.06& 32.43 \\
         &  ZS-BERT & 34.12 &34.38& 34.25& 35.54& 38.19 &36.82 \\
         &  Tran et al. (2022) & 38.37&36.05& 37.17 &43.96 &39.11 &41.36 \\
         &  RelationPrompt NG & 54.45 &29.43 &37.45& 66.49 &40.05 &49.38 \\
         &  RelationPrompt & 63.69 &\underline{67.93}& 65.74& 74.33& 72.51 &73.40 \\
         & RE-Matching & 67.31 & 67.33 & 67.32 & 73.80 & 73.52 & 73.66 \\
         & DSP-ZRSC & \underline{77.5} & 64.4 & \underline{70.4} & \underline{82.9} & 78.1 & \underline{80.4} \\
         &  Tran et al. (2023) &64.68 &65.01& 65.30& 66.44 &69.29 &67.82 \\
         &  MC-BERT & 65.71& 67.11&66.40 &80.71&\underline{79.84}& 80.27 \\
         &  MC-BERT w/ desc (Ours) & 68.53 & 69.81 & 69.16 & 79.22 & 78.19 & 78.69 \\
         &  MC-BERT-KG (Ours) & \textbf{79.28}  &  \textbf{76.95} & \textbf{78.07} & \textbf{84.46} & \textbf{80.90} & \textbf{82.64} \\
         \bottomrule
    \end{tabular}
    \caption{Results on FewRel and Wiki-ZSL}
    \label{tab:zs}
\end{table}

While further inspecting the performance of our method on Re-DocRED, we also saw that the biggest difficulty on the document-level relation extraction datasets was to identify whether any relation is expressed between two entities, not which. Only six percent of errors on Re-DocRED stem from the problem of disambiguating between different relations. We assume that graph information is less helpful to decide whether any relation is expressed as this is more a problem related to the textual relation extraction component. 
 An additional cause might be that the number of relations for both datasets is rather small being fewer than 100. Therefore, the ambiguity between different relations might be rather low.

On BioRel (see Table~\ref{tab:biorel}), our method performs the best.  The textual relation extraction component has again a large impact; however, the inclusion of graph information leads to a performance far beyond the previous SOTA. We suspect that the impact is higher on this dataset as there are more relations to disambiguate in comparison to DWIE and Re-DocRED.

\paragraph{Zero-shot.}

In zero-shot relation extraction, the influence of graph information exhibits variation across tasks. As shown in Table~\ref{tab:zs}, graph information significantly enhances performance in the most challenging scenario ($m=15$) for the Wiki-ZSL dataset. While its impact on FewRel is comparatively modest, our method still surpasses the previous state-of-the-art by two F1-measure points. This indicates that graph information is particularly beneficial when data presents higher complexity or ambiguity, whereas its advantage diminishes in simpler scenarios such as $m=5$ or $m=10$. 
In almost every setting, except for FewRel with $m=5$, incorporating graph information results in a substantial performance enhancement compared to not utilizing it, as evident when comparing to MC-BERT w/ descriptions. It is important to note that integrating graph information is independent of other model improvements, meaning that the current state-of-the-art methods across several datasets could potentially be further enhanced by incorporating this additional information as well.

\subsection{Ablation studies}
In addition to the ablation studies already included in the previous section, we further investigate our method on the $m=15$ split of the Wiki-ZSL dataset. Specifically, we analyse the influence of the number of hops and keeping the links between subject and object entities. 

\begin{table}[]
    \centering
    \begin{tabular}{cccc}
        \toprule
        Model & P & R & F1 \\
        \midrule
         MC-BERT-KG - graph only & 69.63  &  66.16 & 67.78 \\
         MC-BERT-KG - 1 - hop & 70.00  &  71.06 & 70.05 \\
         MC-BERT-KG - 2 - hop & 76.74  &  75.95 & 76.25 \\
         MC-BERT-KG - 3 - hop & 75.44  &  75.46 & 75.44 \\
         MC-BERT-KG - with direct triples & 76.95  &  75.57 & 76.25 \\
         MC-BERT-KG & \textbf{79.28}  &  \textbf{76.95} & \textbf{78.07} \\
         \bottomrule
    \end{tabular}
    \caption{Ablation on Wiki-ZSL $m=15$}
    \label{tab:ablation_wiki}
\end{table}

Analyzing the ablation results for using only the graph model (Table~\ref{tab:ablation_wiki}), we observe competitive performance with other state-of-the-art approaches, indicating that the KG context provides substantial cues to discern the correct relation, especially in scenarios with a clear single correct relation. However, augmenting this with textual information results in a substantial performance increase of approximately 12 F1-measure points. This underscores the complementary nature of text and graph information, highlighting the importance of leveraging both to maximize relation extraction efficacy. One advantage of the Wiki-ZSL dataset is that a relation is consistently expressed between any two entities. If this were not the case, the model would have to rely more heavily on textual context, making the graph information less dependable. This is evident in the significantly poorer performance of KG-only methods on document-level relation extraction datasets, as reflected in Tables~\ref{tab:dwie} and~\ref{tab:rdocred}.

If we use fewer than 4 hops, we see a diminishing performance. 
Surprisingly, only using 2-hops outperforms using at maximum 3-hops of path lengths. As the 1-hop distance is effectively not using any knowledge graph information due to us filtering the single hops out, its performance reduces to the text-only model. Interestingly, including direct triples during training actually reduces performance. We attribute this to the model's tendency to rely on straightforward single-hop information rather than considering the broader context of the surrounding neighborhood, which is not always the optimal approach.

\section{Related Work}


Regular relation extraction is usually tackled as classification problem. The input text is encoded and a classification head is attached. To encode text, CNNs~\cite{Zeng2014}, RNNs~\cite{Miwa2016} or transformers~\cite{Zhong2021} are usually employed. Recently, pre-trained models have been extensively used which are fine-tuned on the relation extraction task~\cite{Zhou2021,Zhang2021}. 

Document-level relation extraction is tackled mostly in two different ways: either by improving the capabilities of pre-trained language models (PLMs) to identify expressed long-range relations~\cite{Zhou2021,Zhang2021} or by representing the text information in a more structured way by either modeling the document as a graph~\cite{Zeng2020,Xu2021} or by learning additional reasoning rules~\cite{DBLP:conf/acl/QiDW24,Fan2022,Ru2021}. Our method is connected to both as we combine the SOTA-performing models relying on PLMs with the graph representation using graph neural networks.

In regard to zero-shot relation extraction usually  representation-learning-based methods~\cite{tran2023enhancing,Tran2022,Chen2021,Zhao2023} try to embed the textual information and the relational information in the same vector space. The relational information such as labels or descriptions is transformed into a representation of the relation. Representations are learned such that the representation of the true relation resides close to a representation of the text in the vector space while the false relation representations are pushed further away. 
Recently, generative language models have been increasingly utilized for the task~\cite{Ni2022,Chen2022,Chia2022,Lv2023}.  Here, the model is prompted with the input text as well as information on the potential relations. The model is then fine-tuned to either generate the relation as expressed in the input text or a full triple consisting of the two entities and relation.
Some methods model the problem as a textual entailment problem~\cite{Rahimi2023,Sainz2021,Obamuyide2018}. 
The method by Lan et al.~\cite{Lan2023} models relation classification as a multiple-choice problem where the text is encoded with relation information and a score is calculated. This is done for all relations and the relation with the highest score is taken.  As this method can be equipped with knowledge graph information, we extend this method in our zero-shot relation extraction experiments.

Relation extraction under the use of knowledge graph information is an underexplored area. In recent times, there have been only a few methods investigating this problem. Most methods either incorporate only one-hop information of entities or rely on trained static representations of entities making the generalization to unseen entities difficult~\cite{Bastos2021,Jain2024,Verlinden2021,Vashishth2018,moller2024incorporating}. Others linearize the underlying graph information while not considering the structural information~\cite{Jain2023}. 
In contrast, our method can generalize to new entities and even relations while considering multi-hop information.

\section{Conclusion and Future Work}
We showed that the inclusion of graph information leads to consistent improvements when doing relation extraction. The improvements are especially noticeable when the textual components of the relation extraction method are not well-trained on the relations to be extracted due to a lack of data. On datasets where the main challenge is to identify whether any relation is holding, the benefit of graph information reduces. This applies to document-level relation extraction datasets.
Due to that, the inclusion of such information is therefore especially worthwhile in the zero-shot setting.

Incorporating a post-prediction step by including the relations predicted via a textual component proved to be beneficial in document-level relation extraction. This is especially the case when confronted with very long documents.

In future works, we want to improve the efficiency of the inclusion of graph information. As the subgraphs are sampled in a very simple way a lot of nodes are introduced which are not of relevance to the task of identifying the relation. Introducing a more sensible sampling strategy would alleviate that problem. Additionally, scalability poses a challenge in the zero-shot setting, as the graph neural networks must be executed separately for each relation, leading to substantial computational overhead. Addressing this limitation is a critical challenge, not only for relation extraction tasks but also for link prediction in general. Finally, the method assumes the existence of a path between two entities. While this is a reasonable assumption in the link prediction task, when additional textual evidence is available using the information in the graph not directly related to an existing path might be valuable as well. 

We omitted LLMs in this paper to be comparable to existing state-of-the-art methods. Furthermore, our investigation is orthogonal to potential improvements by using a fine-tuned LLM as they can also be equipped with the graph component to consider knowledge graph information as well. 

\begin{credits}

\subsubsection{\ackname} 
This project was supported by the Hub of Computing and Data Science (HCDS) of Hamburg University within the Cross-Disciplinary Lab program. Additionally, support was provided by the Ministry of Research and Education within the SifoLIFE project "RESCUE-MATE: Dynamische Lageerstellung und Unterstützung für Rettungskräfte in komplexen Krisensituationen mittels Datenfusion und intelligenten Drohnenschwärmen" (FKZ 13N16836).
\end{credits}

\bibliographystyle{plain} 
\bibliography{main}

\end{document}